\documentclass[runningheads]{llncs}

\usepackage{eccv}

\usepackage{pifont} 
\usepackage{multirow}
\usepackage{amsmath}
\usepackage{circledsteps}
\usepackage{bm}
\usepackage{eccvabbrv}

\usepackage{graphicx}
\usepackage{booktabs}

\usepackage[accsupp]{axessibility}  

\usepackage{hyperref}

\usepackage{orcidlink}

\begin{document}

\title{OOTDiffusion: Outfitting Fusion based Latent Diffusion for Controllable Virtual Try-on}

\titlerunning{OOTDiffusion for Controllable Virtual Try-on}

\author{Yuhao Xu \and
Tao Gu \and
Weifeng Chen \and
Chengcai Chen}

\authorrunning{Xu et al.}

\institute{Xiao-i Research\\
\email{\{yuhao.xu,tao.gu,weifeng.chen,arlenecc\}@xiaoi.com}}

\maketitle

\begin{figure}[!h]
  \centering
  \includegraphics[width=\linewidth]{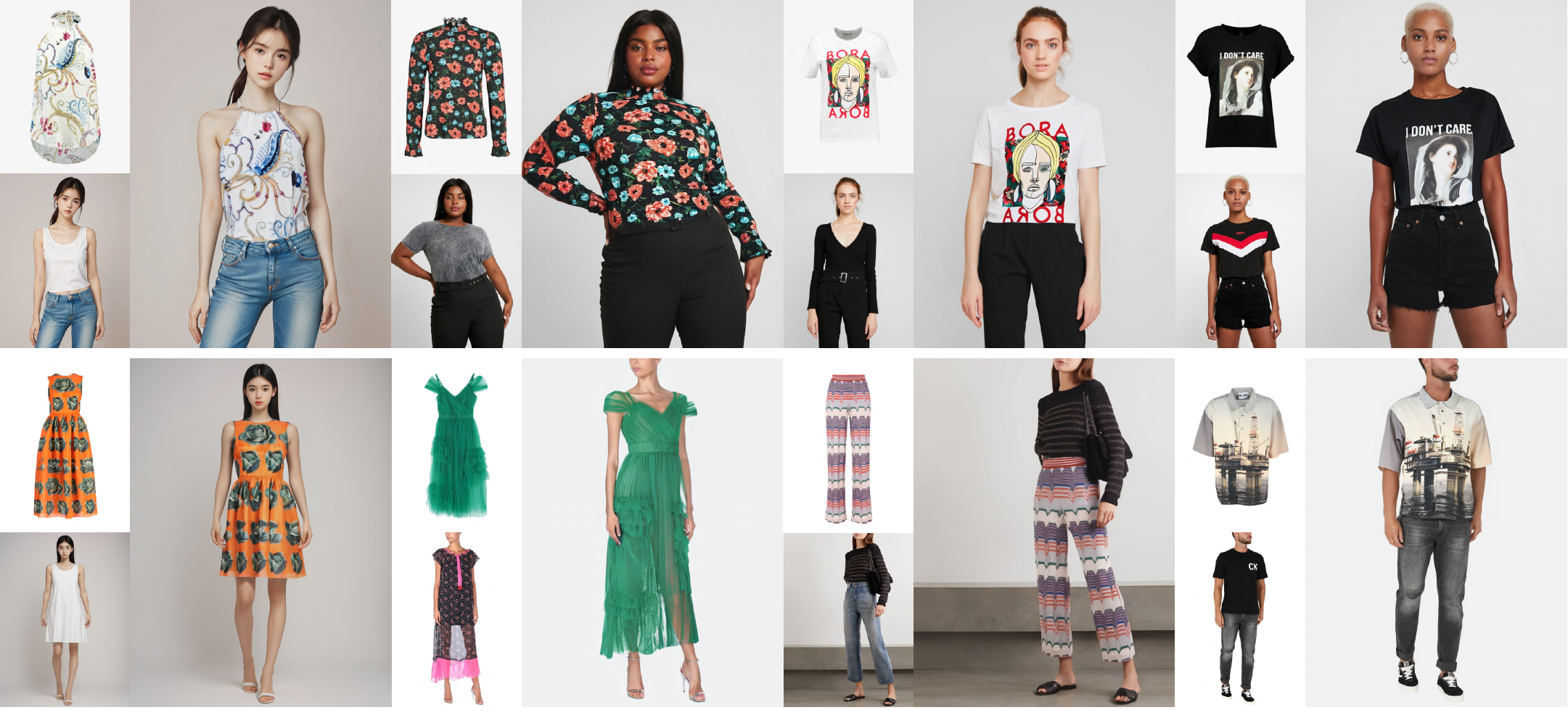}
  \caption{Outfitted images ($1024\times 768$) generated by our OOTDiffusion trained on the VITON-HD~\cite{choi2021viton} (1st row; supporting upper-body garments) and Dress Code~\cite{morelli2022dress} (2nd row; supporting upper-body garments, lower-body garments and dresses) datasets, with various input human and garment images. Please zoom in for more details.}
  \label{fig:demo}
  \vspace{-0.5cm}
\end{figure}

\begin{abstract}
We present OOTDiffusion, a novel network architecture for realistic and controllable image-based virtual try-on (VTON). We leverage the power of pretrained latent diffusion models, designing an outfitting UNet to learn the garment detail features. Without a redundant warping process, the garment features are precisely aligned with the target human body via the proposed outfitting fusion in the self-attention layers of the denoising UNet. In order to further enhance the controllability, we introduce outfitting dropout to the training process, which enables us to adjust the strength of the garment features through classifier-free guidance. Our comprehensive experiments on the VITON-HD and Dress Code datasets demonstrate that OOTDiffusion efficiently generates high-quality try-on results for arbitrary human and garment images, which outperforms other VTON methods in both realism and controllability, indicating an impressive breakthrough in virtual try-on.
Our source code is available at \url{https://github.com/levihsu/OOTDiffusion}.
\end{abstract}

\section{Introduction}
\label{sec:intro}
Image-based virtual try-on (VTON) is a popular and promising image-synthesis technology for the current e-commerce industry, which is able to dramatically improve the shopping experience of consumers and reduce the advertising cost of clothing merchants. As its name suggests, the VTON task aims to generate an outfitted image of a target human wearing a given garment. It has taken tremendous efforts from numerous researchers~\cite{han2018viton, wang2018toward, han2019clothflow, dong2019towards, minar2020cp, yang2020towards, ge2021parser, ge2021disentangled, he2022style, xie2023gp, kim2023stableviton} for more natural and accurate virtual try-on results in the past few years.

Image-based VTON is currently facing two main challenges. First, the generated images should be realistic and natural enough to avoid dissonance. Most of recent researches on virtual try-on leverage generative adversarial networks~\cite{goodfellow2020generative} (GANs) or latent diffusion models~\cite{rombach2022high} (LDMs) for image generation. Previous GAN-based methods~\cite{han2018viton, han2019clothflow, choi2021viton, he2022style, lee2022high, xie2023gp} usually have difficulty in generating correct garment folds, natural light and shadow, or realistic human bodies. Hence more recent work favors LDM-based methods~\cite{zhu2023tryondiffusion, morelli2023ladi, gou2023taming, kim2023stableviton}, which effectively improve the realism of outfitted images. The second critical challenge is how to preserve as much as possible the garment detail features, such as complicated text, textures, colors, patterns and lines, etc. Previous researches~\cite{choi2021viton, lee2022high, morelli2023ladi, xie2023gp, gou2023taming} usually perform an explicit warping process to align the garment features with the target human body, and then feed the warped garment into generative models (i.e., GANs and LDMs, etc.). Thus the performance of such VTON methods is extremely dependent on the efficacy of the independent warping process which is prone to overfitting the training data. On the other hand, some LDM-based methods~\cite{morelli2023ladi, gou2023taming, kim2023stableviton} attempt to learn garment features via CLIP textual-inversion~\cite{gal2022image}, which fail to preserve fine-grained garment details.

Motivated by the aforementioned prospects and challenges of image-based VTON, we present a novel LDM-based virtual try-on method, namely Outfitting over Try-on Diffusion (OOTDiffusion; see \cref{fig:workflow}). First, we make full use of the advantages of pretrained latent diffusion models~\cite{rombach2022high} to ensure high realism of generated images and natural try-on effects, and design an outfitting UNet to learn the detail features of garments in the latent space in a single step. Then, we propose an outfitting fusion process to precisely align the garment features with the noisy human body in the self-attention layers~\cite{vaswani2017attention} of the denoising UNet. In this way, the garment features are smoothly adapted to various target human body types and postures, without suffering information loss or feature distortion caused by an independent warping process. Furthermore, we perform an outfitting dropout operation, randomly dropping a handful of garment latents in training to enable classifier-free guidance~\cite{ho2022classifier} with respect to the garment features. Through this approach, the strength of garment control over the generated result can be simply adjusted by a guidance scale, which further enhances the controllability of our VTON method.

Our contributions are summarized as follows:
\begin{itemize}
\item We present OOTDiffusion, an LDM-based network architecture with a novel outfitting UNet for realistic and controllable virtual try-on.
\item We propose outfitting fusion to efficiently align the garment features with the target human body in the self-attention layers without redundant warping.
\item We introduce outfitting dropout to the training process, which further improves the controllability of the outfitting UNet.
\item We train our OOTDiffusion on two broadly-used high-resolution benchmark datasets, i.e., VITON-HD~\cite{choi2021viton} and Dress Code~\cite{morelli2022dress}, respectively. Extensive qualitative and quantitative evaluations demonstrate our superiority over the state-of-the-art VTON methods in both realism and controllability for various target human and garment images (see \cref{fig:demo}), implying an impressive breakthrough in image-based virtual try-on.
\end{itemize}

\section{Related Work}

\subsubsection{Image-based Virtual Try-on.}
Image-based virtual try-on has been investigated for many years as a promising and challenging task~\cite{zhu2023tryondiffusion,wang2018toward,issenhuth2020not,han2018viton,fenocchi2022dual,gou2023taming,morelli2022dress,xie2023gp,kim2023stableviton,lee2022high,morelli2023ladi,choi2021viton}. Aiming at more natural and accurate results, recent researches are mainly based on generative adversarial networks~\cite{goodfellow2020generative} (GANs) or latent diffusion models~\cite{rombach2022high} (LDMs) for image generation. Among the GAN-based VTON methods~\cite{choi2021viton,lee2022high,xie2023gp}, VITON-HD~\cite{choi2021viton} collected a high-resolution dataset and proposed ALIAS normalization and generator to address the misalignment between warped clothes and target regions. HR-VITON~\cite{lee2022high} simultaneously performed warping and segmentation to handle the body occlusion and garment misalignment. GP-VTON~\cite{xie2023gp} proposed an LFGP warping module to generate deformed garments and introduced a DGT training strategy for the warping network. As introduced above, GAN-based methods usually rely on an explicit warping process neglecting realistic garment folds and natural light and shadow, which seriously degrades the fidelity and realism of outfitted images. Meanwhile, GAN-based methods are prone to overfitting the training data and causing severe performance degradation on out-of-distribution images.

With respect to the LDM-based approaches~\cite{morelli2023ladi,gou2023taming,kim2023stableviton}, LaDI-VTON~\cite{morelli2023ladi} and DCI-VTON~\cite{gou2023taming} also require an explicit warping process. In specific, LaDI-VTON~\cite{morelli2023ladi} performed textual-inversion to map the visual garment features to the CLIP~\cite{radford2021learning} token embedding space and condition the latent diffusion model along with the warped input. DCI-VTON~\cite{gou2023taming} directly combined the warped clothes with the masked person image to get a coarse result, and then refined it by the diffusion model. Neither of these methods succeeded in fully preserving garment details like complicated patterns and text due to the information loss caused by the CLIP encoder. More recently, StableVITON~\cite{kim2023stableviton} discarded independent warping and proposed a zero cross-attention block to learn semantic correlation between the clothes and human body. However, information loss remains in the cross-attention layers, and the extra zero-initialized blocks heavily increase the training and inference cost. In contrast, again without warping, our LDM-based OOTDiffusion finetunes the pretrained outfitting UNet to learn garment details in one step and efficiently incorporates them into the denoising UNet via our outfitting fusion with negligible information loss. 

\subsubsection{LDM-based Controllable Image Generation.} Latent diffusion models~\cite{rombach2022high} have achieved great success in text-to-image~\cite{podell2023sdxl,betker2023improving,saharia2022photorealistic,ruiz2023dreambooth,kumari2023multi} and image-to-image~\cite{saharia2022palette,kawar2023imagic,saharia2022image,tumanyan2023plug,parmar2023zero} generation in recent years. For the purpose of more controllable generated results, Prompt-to-Prompt~\cite{hertz2022prompt} and Null-text Inversion~\cite{mokady2023null} controlled the cross-attention layers to finely edit images by modifying the input captions without extra model training. InstructPix2Pix~\cite{brooks2023instructpix2pix} created paired data to train diffusion models that generate the edited image given an input image and a text instruction. Paint-by-Example~\cite{yang2023paint} trained image-conditioned diffusion models in a self-supervised manner to offer fine-grained image control. ControlNet~\cite{zhang2023adding} and T2I-Adapter~\cite{mou2023t2i} incorporated additional blocks into pretrained diffusion models to enable spatial conditioning controls. IP-Adapter~\cite{ye2023ip} adopted a decoupled cross-attention mechanism for text and image features to enable controllable generation with image prompt and additional structural conditions. In this paper, we focus on the image-based VTON task, employing outfitting fusion in the self-attention layers of the denoising UNet and performing outfitting dropout at training time to enable latent diffusion models to generate more controllable outfitted images with respect to the garment features.

\section{Method}

\subsection{Preliminary}

\subsubsection{Stable Diffusion.}
Our OOTDiffusion is an extension of Stable Diffusion~\cite{rombach2022high}, which is one of the most commonly-used latent diffusion models. Stable Diffusion employs a variational autoencoder~\cite{kingma2013auto} (VAE) that consists of an encoder $\mathcal{E}$ and a decoder $\mathcal{D}$ to enable image representations in the latent space. And a UNet~\cite{ronneberger2015u} $\epsilon_{\theta}$ is trained to denoise a Gaussian noise $\epsilon$ with a conditioning input encoded by a CLIP text encoder~\cite{radford2021learning} $\tau_{\theta}$. Given an image $\mathbf{x}$ and a text prompt $\mathbf{y}$, the training of the denoising UNet $\epsilon_{\theta}$ is performed by minimizing the following loss function:
\begin{equation}
    \mathcal{L}_{LDM} = \mathbb{E}_{\mathcal{E}(\mathbf{x}),\mathbf{y},\epsilon\sim\mathcal{N}(0, 1),t}\left[\lVert\epsilon - \epsilon_{\theta}(\mathbf{z}_t, t, \tau_{\theta}(\mathbf{y}))\rVert_2^2\right],
\end{equation}
where $t\in\{1,...,T\}$ denotes the time step of the forward diffusion process, and $\mathbf{z}_t$ is the encoded image $\mathcal{E}(\mathbf{x})$ with the added Gaussian noise $\epsilon\sim\mathcal{N}(0, 1)$ (i.e., the noise latent). Note that the conditioning input $\tau_{\theta}(\mathbf{y})$ is correlated with the denoising UNet by the cross-attention mechanism~\cite{vaswani2017attention}.

\subsection{OOTDiffusion}

\subsubsection{Overview.}

\begin{figure}[!t]
  \centering
  \includegraphics[width=\linewidth]{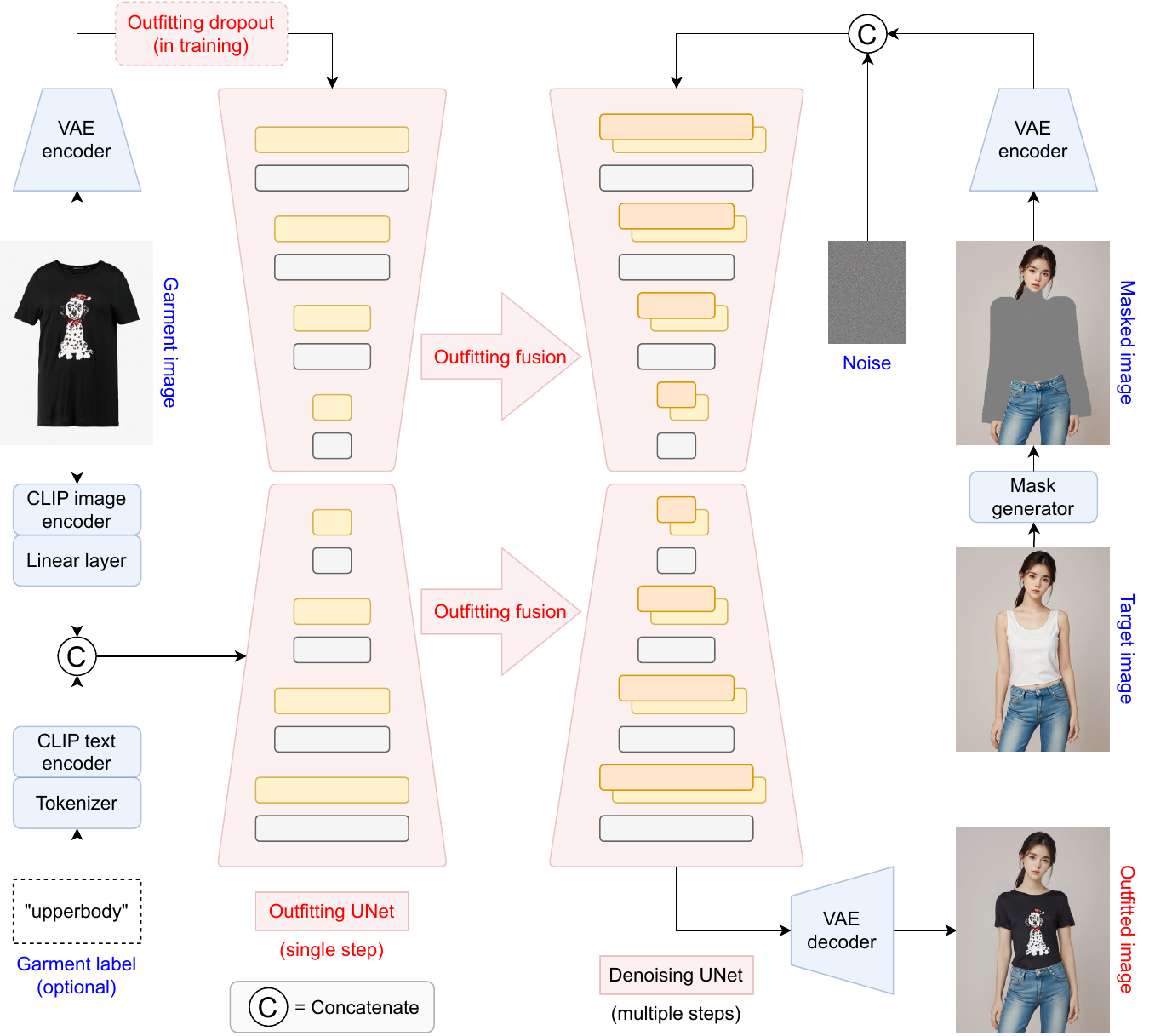}
  \caption{Overview of our proposed OOTDiffusion model. On the left side, the garment image is encoded into the latent space and fed into the outfitting UNet for a single step process. Along with the auxiliary conditioning input generated by CLIP encoders, the garment features are incorporated into the denoising UNet via outfitting fusion. Outfitting dropout is performed for the garment latents particularly in training to enable classifier-free guidance. On the right side, the input human image is masked with respect to the target region and concatenated with a Gaussian noise as the input to the denoising UNet for multiple sampling steps. After denoising, the feature map is decoded back into the image space as our try-on result.}
  \label{fig:workflow}
  \vspace{-0.5cm}
\end{figure}

\cref{fig:workflow} illustrates the overview of our method. Given a target human image $\mathbf{x}\in\mathbb{R}^{3\times H\times W}$ and an input garment image $\mathbf{g}\in\mathbb{R}^{3\times H\times W}$, OOTDiffusion is capable of generating a realistic outfitted image $\mathbf{x_g}\in\mathbb{R}^{3\times H\times W}$. We employ OpenPose~\cite{8765346,simon2017hand,cao2017realtime,wei2016cpm} and HumanParsing~\cite{li2020self} to generate a masked human image $\mathbf{x_m}\in\mathbb{R}^{3\times H\times W}$, and use a VAE encoder $\mathcal{E}$ to transform it into the latent space as $\mathcal{E}(\mathbf{x_m})\in\mathbb{R}^{4\times h\times w}$, where $h=\frac{H}{8}$ and $w=\frac{W}{8}$. Then we concatenate $\mathcal{E}(\mathbf{x_m})$ with a Gaussian noise $\epsilon\in\mathbb{R}^{4\times h\times w}$ as the input latent $\mathbf{z}_T\in\mathbb{R}^{8\times h\times w}$ for the denoising UNet. Note that we add $4$ zero-initialized channels to the first convolutional layer of the denoising UNet to support our input with $8$ channels.

On the other side, we feed the encoded garment latent $\mathcal{E}(\mathbf{g})\in\mathbb{R}^{4\times h\times w}$ into an (i) outfitting UNet to learn the garment features in a single step, and integrate them into the denoising UNet via our (ii) outfitting fusion. And we perform (iii) outfitting dropout for $\mathcal{E}(\mathbf{g})$ particularly in the training process. In addition, we also conduct CLIP textual-inversion~\cite{gal2022image} for the garment image $\mathbf{g}$, and optionally concatenate it with a text embedding of the garment label $\mathbf{y}\in\{``upperbody",``lowerbody",``dress"\}$ as an auxiliary conditioning input, which is fed into both outfitting and denoising UNets via the cross-attention mechanism~\cite{vaswani2017attention}. Finally, after multiple steps of the denoising process, we use a VAE decoder $\mathcal{D}$ to transform the denoised latent $\mathbf{z}_0\in\mathbb{R}^{4\times h\times w}$ back into the image space as the output image $\mathbf{x_g}=\mathcal{D}(\mathbf{z}_0)\in\mathbb{R}^{3\times H\times W}$. We will elaborate the key technologies (i.e., (i) outfitting UNet, (ii) outfitting fusion, and (iii) outfitting dropout) of our OOTDiffusion in the following sections.

\subsubsection{Outfitting UNet.}
As introduced above, we propose an outfitting UNet to efficiently learn the detail features of the garment image $\mathbf{g}$. The left side of \cref{fig:workflow} shows the architecture of our outfitting UNet, which is essentially identical to the denoising UNet of Stable Diffusion. The encoded garment latent $\mathcal{E}(\mathbf{g})\in\mathbb{R}^{4\times h\times w}$ is fed into the outfitting UNet $\omega_{\theta'}$, and then incoporated into the denoising UNet $\epsilon_{\theta}$ via our outfitting fusion (see the next section). Along with the aforementioned auxiliary conditioning input, the outfitting and denoising UNets are jointly trained by minimizing the following loss function:
\begin{equation}
\label{eq:loss}
    \mathcal{L}_{OOTD} = \mathbb{E}_{\mathcal{E}(\mathbf{x_m}),\mathcal{E}(\mathbf{g}),\psi,\epsilon\sim\mathcal{N}(0, 1),t}\left[\lVert\epsilon - \epsilon_{\theta}(\mathbf{z}_t,t,\omega_{\theta'}(\mathcal{E}(\mathbf{g}),\psi),\psi)\rVert_2^2\right],
\end{equation}
where $\psi=\tau_g(\mathbf{g})\ \textcircled{c}\ \tau_y(\mathbf{y})$ represents the auxiliary conditioning input for both $\omega_{\theta'}$ and $\epsilon_{\theta}$. While $\tau_g$ and $\tau_y$ refer to the pretrained CLIP image encoder and text encoder respectively, and \textcircled{c} denotes concatenation.

In practice, we directly duplicate the pretrained UNet weights of Stable Diffusion~\cite{rombach2022high} for the initialization of both our outfitting and denoising UNets (except for the zero-initialized channels added to the first convolutional layer), and jointly finetune them on the high-resolution VTON datasets~\cite{choi2021viton,morelli2022dress}. Note that $\omega_{\theta'}$ and $\epsilon_{\theta}$ do not share any weights in the training process. We claim that our tactical utilization of the pretrained models dramatically improves the training efficiency and reduces the training cost.
Moreover, compared with the denoising UNet, a significant difference in our outfitting UNet is that it requires only one step forward process before the multiple denoising steps in inference, causing a minimal amount of extra computational cost to the original Stable Diffusion~\cite{rombach2022high}.

\subsubsection{Outfitting Fusion.}

\begin{figure}[!t]
  \centering
  \includegraphics[width=\linewidth]{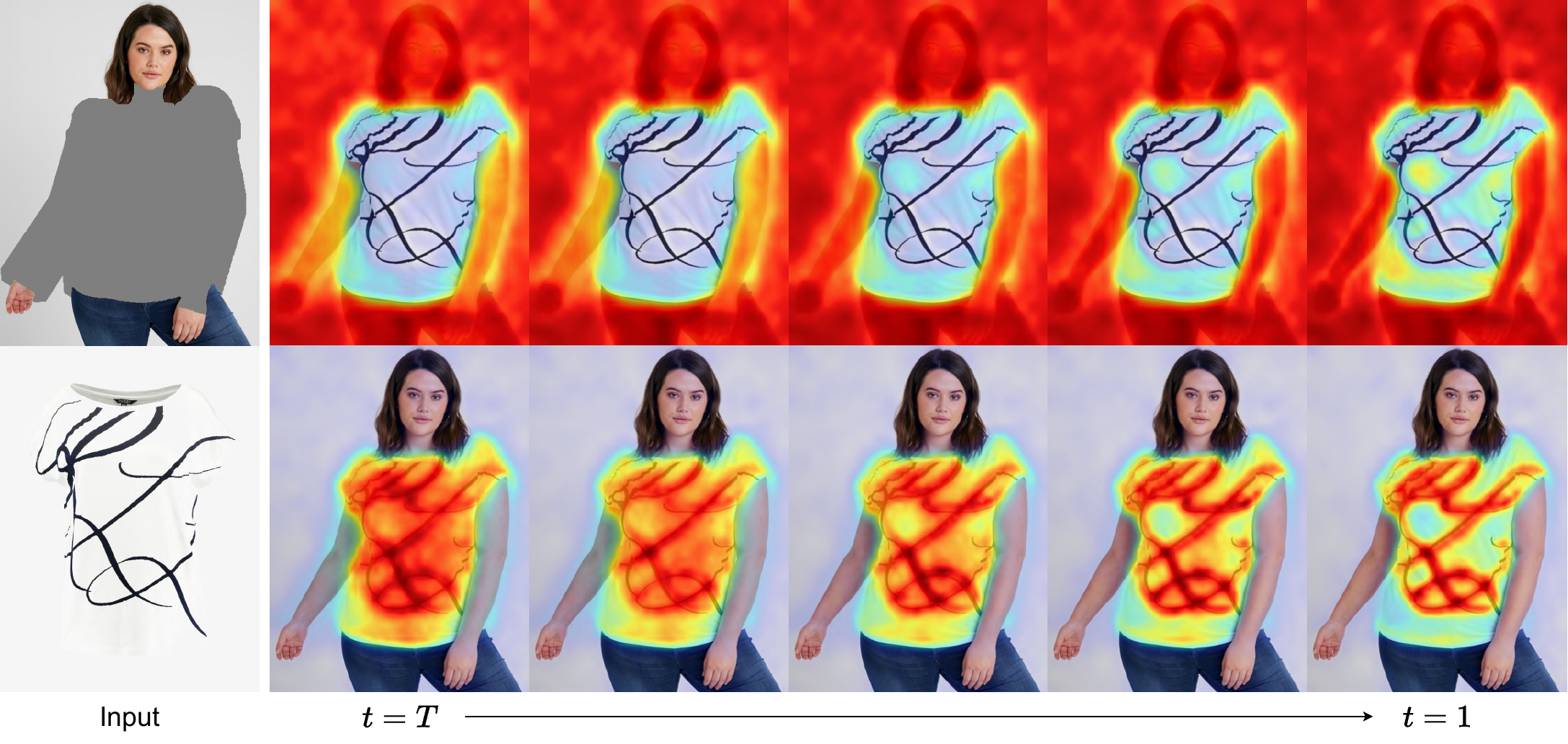}
  \caption{Visualization of the attention maps with respect to the human body (1st row) and garment features (2nd row) aligned by our outfitting fusion.}
  \label{fig:fusion}
  \vspace{-0.5cm}
\end{figure}

Based on our proposed outfitting UNet and inspired by the spatial-attention mechanism~\cite{vaswani2017attention,hu2023animate}, we propose an outfitting fusion process to incorporate the learned garment features into the denoising UNet. First, we dive into the transformer blocks~\cite{vaswani2017attention} of two UNets, finding each pair of feature maps used as the input to the corresponding self-attention layers~\cite{vaswani2017attention}. Given the $n$th pair of the feature maps $\mathbf{g}_n,\mathbf{x}_n\in\mathbb{R}^{c_n\times h_n\times w_n}$, we concatenate them in the spatial domain as:
\begin{equation}
    \mathbf{x_g}_n=\mathbf{x}_n\ \textcircled{c}\ \mathbf{g}_n\in\mathbb{R}^{c_n\times h_n\times 2w_n}.
\end{equation}
And we replace $\mathbf{x}_n$ with the concatenated feature map $\mathbf{x_g}_n$ as the input to the self-attention layer of the denoising UNet. Then we crop out the fist half of the output feature map as the final output of the self-attention layer. \cref{fig:fusion} visualizes the attention maps learned in our modified self-attention. We observe that the unmasked region focuses attention on the human body (1st row), and the masked pixels are attracted to the garment features (2nd row). Meanwhile, during the denoising process, the attention to the human body gradually includes part of the masked region like the neck and arms, and the attention to the garment features gradually increases in the region of the complicated patterns. Through outfitting fusion in the self-attention layers, the garment features are implicitly warped and effectively correlated with the target human body with negligible information loss. Hence the denoising UNet is made capable of learning the precise features from the outfitting UNet for preserving garment details and naturally adapting them to the target human body in the generated image.

\subsubsection{Outfitting Dropout.}
In order to further enhance the controllability of our VTON method, we employ an outfitting dropout operation in training to enable classifier-free guidance~\cite{ho2022classifier} with respect to the garment features. Classifier-free guidance has been broadly used in conditional image generation~\cite{nichol2021glide,yu2022scaling,saharia2022photorealistic,brooks2023instructpix2pix} for trading off the quality and diversity of images generated by latent diffusion models. Specifically in the training process of our outfitting UNet, we randomly drop the input garment latent as $\mathcal{E}(\mathbf{g})=\varnothing$, where $\varnothing\in\mathbb{R}^{4\times h\times w}$ refers to an all-zero latent. In this way, the denoising UNet is trained both conditionally and unconditionally, i.e., with and without the outfitting fusion. Then at inference time, we simply use a guidance scale $s_\mathbf{g}\geq 1$ to adjust the strength of conditional control over the predicted noise $\hat{\epsilon}_{\theta}$ as:
\begin{equation}
\label{eq:sg}
    \hat{\epsilon}_{\theta}(\mathbf{z}_t,\omega_{\theta'}(\mathcal{E}(\mathbf{g})))=\epsilon_{\theta}(\mathbf{z}_t,\varnothing)+s_\mathbf{g}\cdot(\epsilon_{\theta}(\mathbf{z}_t,\omega_{\theta'}(\mathcal{E}(\mathbf{g})))-\epsilon_{\theta}(\mathbf{z}_t,\varnothing)),
\end{equation}
where we omit some minor terms compared with \cref{eq:loss} for the sake of brevity.

In practice, we empirically set the outfitting dropout ratio to $10\%$ in training, i.e., $10\%$ of garment latents $\mathcal{E}(\mathbf{g})$ are set to $\varnothing$. And the optimal value of the guidance scale $s_\mathbf{g}$ is usually around $1.5\sim 2.0$ according to our ablation study (see \cref{sec:ablat}). \cref{fig:ablation} and \cref{tab:abl_component} demonstrate the effects of our outfitting dropout and different guidance scale values.

\section{Experiments}
\label{sec:exper}

\subsection{Experimental Setup}

\subsubsection{Datasets.}
Our experiments are performed on two high-resolution ($1024\times 768$) virtual try-on datasets, i.e., VITON-HD~\cite{choi2021viton} and Dress Code~\cite{morelli2022dress}. The VITON-HD dataset consists of 13,679 image pairs of frontal half-body models and corresponding upper-body garments, where 2032 pairs are used as the test set. The Dress Code dataset consists of 15,363/8,951/2,947 image pairs of full-body models and corresponding upper-body garments/lower-body garments/dresses, where 1,800 pairs for each garment category are used as the test set.

\subsubsection{Compared Methods.}
On the VITON-HD dataset~\cite{choi2021viton}, we compare our OOTDiffusion with multiple state-of-the-art VTON methods, including the GAN-based VITON-HD~\cite{choi2021viton}, HR-VITON~\cite{lee2022high} and GP-VTON~\cite{xie2023gp}, as well as the LDM-based LaDI-VTON~\cite{morelli2023ladi} and StableVITON~\cite{kim2023stableviton}.

While for the evaluation on the Dress Code dataset~\cite{morelli2022dress}, since VITON-HD~\cite{choi2021viton}, HR-VITON~\cite{lee2022high} and StableVITON~\cite{kim2023stableviton} are not designed for the entire dataset beyond upper-body garments, we select two VTON methods (i.e., GP-VTON~\cite{xie2023gp} and LaDI-VTON~\cite{morelli2023ladi}) and another LDM-based inpainting method (i.e., Paint-by-Example~\cite{yang2023paint}) for fair comparison. 

\subsubsection{Evaluation Metrics.}
We evaluate the results in both the paired and unpaired settings, where the paired setting provides the target human and the corresponding garment images for reconstruction, and the unpaired setting provides the different garment images for virtual try-on. Specifically for Dress Code~\cite{morelli2022dress}, we note that the evaluation is performed on the entire dataset rather than being limited to upper-body garments. This more effectively validates the feasibility of each method in real-world applications with various garment types. 

In the quantitative evaluation, though our OOTDiffusion supports higher-resolution ($1024\times 768$) virtual try-on, all the experiments are conducted at the resolution of $512\times 384$ for fair comparison with previous VTON methods. For the paired setting, we use LPIPS~\cite{zhang2018unreasonable} and SSIM~\cite{wang2004image} to measure the quality of the generated image in terms of restoring the original image. For the unpaired setting, we employ FID~\cite{heusel2017gans} and KID~\cite{binkowski2018demystifying} for realism and fidelity assessment. We follow the previous work~\cite{detlefsen2022torchmetrics,parmar2022aliased,morelli2023ladi} to implement all of these metrics.

\subsection{Implementation Details}
\label{sec:imple}
In our experiments, we initialize the OOTDiffusion models by inheriting the pretrained weights of Stable Diffusion v1.5~\cite{rombach2022high}. Then we finetune the outfitting and denoising UNets using an AdamW optimizer~\cite{loshchilov2018fixing} with a fixed learning rate of 5e-5. Note that we train four types of models on VITON-HD~\cite{choi2021viton} and Dress Code~\cite{morelli2022dress} datasets at resolutions of $512\times 384$ and $1024\times 768$, separately. All the models are trained for 36,000 iterations on a single NVIDIA A100 GPU, with a batch size of 64 for the $512\times 384$ resolution and 16 for the $1024\times 768$ resolution. At inference time, we run our OOTDiffusion on a single NVIDIA RTX 4090 GPU for 20 sampling steps using the UniPC sampler~\cite{zhao2024unipc}.

\subsection{Ablation Study}
\label{sec:ablat}

\begin{figure}[!t]
  \centering
  \includegraphics[width=\linewidth]{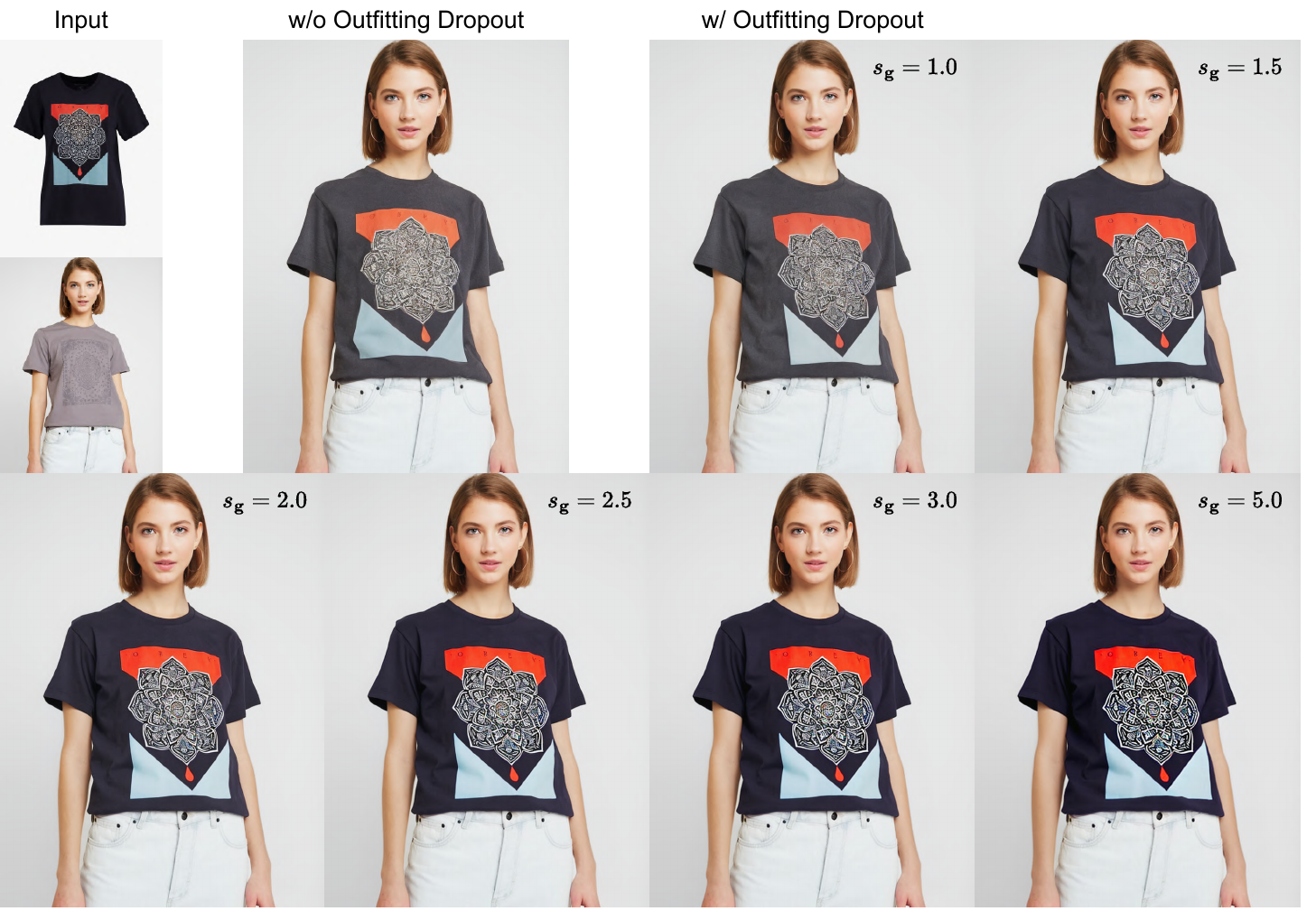}
  \caption{Qualitative comparison of outfitted images generated by OOTDiffusion models trained without/with outfitting dropout and using different values of the guidance scale $s_\mathbf{g}$. Please zoom in for more details.}
  \label{fig:ablation}
\end{figure}

\begin{table}[!t]
\caption{Ablation study of outfitting dropout and different guidance scale values on the VITON-HD dataset~\cite{choi2021viton}. The best and second best results are reported in \textbf{bold} and \underline{underline}, respectively.}
\centering
\begin{tabular}{cc|cccc} 
\toprule
\textbf{Outfitting} &\textbf{Guidance} & \multirow{2}{*}{\textbf{LPIPS} $\downarrow$} & \multirow{2}{*}{\textbf{SSIM} $\uparrow$} & \multirow{2}{*}{\textbf{FID} $\downarrow$} & \multirow{2}{*}{\textbf{KID} $\downarrow$} \\ 
\textbf{Dropout} &\textbf{Scale} & & & &\\
\hline
\ding{55} & -   & 0.0750                  & 0.8699               & 8.91                 & 0.89          \\
\ding{51} & 1.0 & 0.0749                  & 0.8705               & 8.99                 & 0.89          \\
\ding{51} & 1.5 & \textbf{0.0705}         & \textbf{0.8775}      & \underline{8.81}     & \textbf{0.82} \\
\ding{51} & 2.0 & \underline{0.0708}      & \underline{0.8766}   & \textbf{8.80}        & \underline{0.86} \\
\ding{51} & 2.5 & 0.0746                  & 0.8691               & 8.84                 & 0.89          \\
\ding{51} & 3.0 & 0.0753                  & 0.8684               & 8.95                 & 0.96          \\
\ding{51} & 5.0 & 0.0788                  & 0.8640               & 9.28                 & 1.22          \\
\bottomrule
\end{tabular}
\label{tab:abl_component}
\vspace{-0.25cm}
\end{table}

We investigate the effects of our proposed outfitting dropout as well as the different values of the guidance scale $s_\mathbf{g}$ on the VITON-HD dataset~\cite{choi2021viton}. First, we train two variants of our OOTDiffusion models without/with outfitting dropout, respectively. Then for the model trained with outfitting dropout, we set $s_\mathbf{g}=1.0,1.5,2.0,2.5,3.0,5.0$ for classifier-free guidance. At inference time, we guarantee all of other parameters (including the random seed) are consistent for fair comparison. As \cref{fig:ablation} shows, without outfitting dropout, classifier-free guidance is not supported and the generated result is obviously the worst. While for the model trained with outfitting dropout, when $s_\mathbf{g}=1.0$, the inference process is identical to the model without outfitting dropout (see \cref{eq:sg}), which gets a similarly bad result. When $s_\mathbf{g}>1.0$, we see that the fine-grained garment features become clearer as $s_\mathbf{g}$ increases. However, color distortion occurs when $s_\mathbf{g}\geq 2.5$ and becomes extremely significant when $s_\mathbf{g}=5.0$ (see the flower patterns). Furthermore, \cref{tab:abl_component} quantitatively proves the efficacy of our outfitting dropout which enables classifier-free guidance with respect to the garment features, and finds the optimal guidance scale value is around $1.5\sim 2.0$ in most cases. According to this study, we consistently conduct outfitting dropout for OOTDiffusion, and empirically set $s_\mathbf{g}=1.5$ for the VITON-HD dataset~\cite{choi2021viton} and $s_\mathbf{g}=2.0$ for the Dress Code dataset~\cite{morelli2022dress} in the following experiments.

\begin{figure}[!t]
  \centering
  \includegraphics[width=\linewidth]{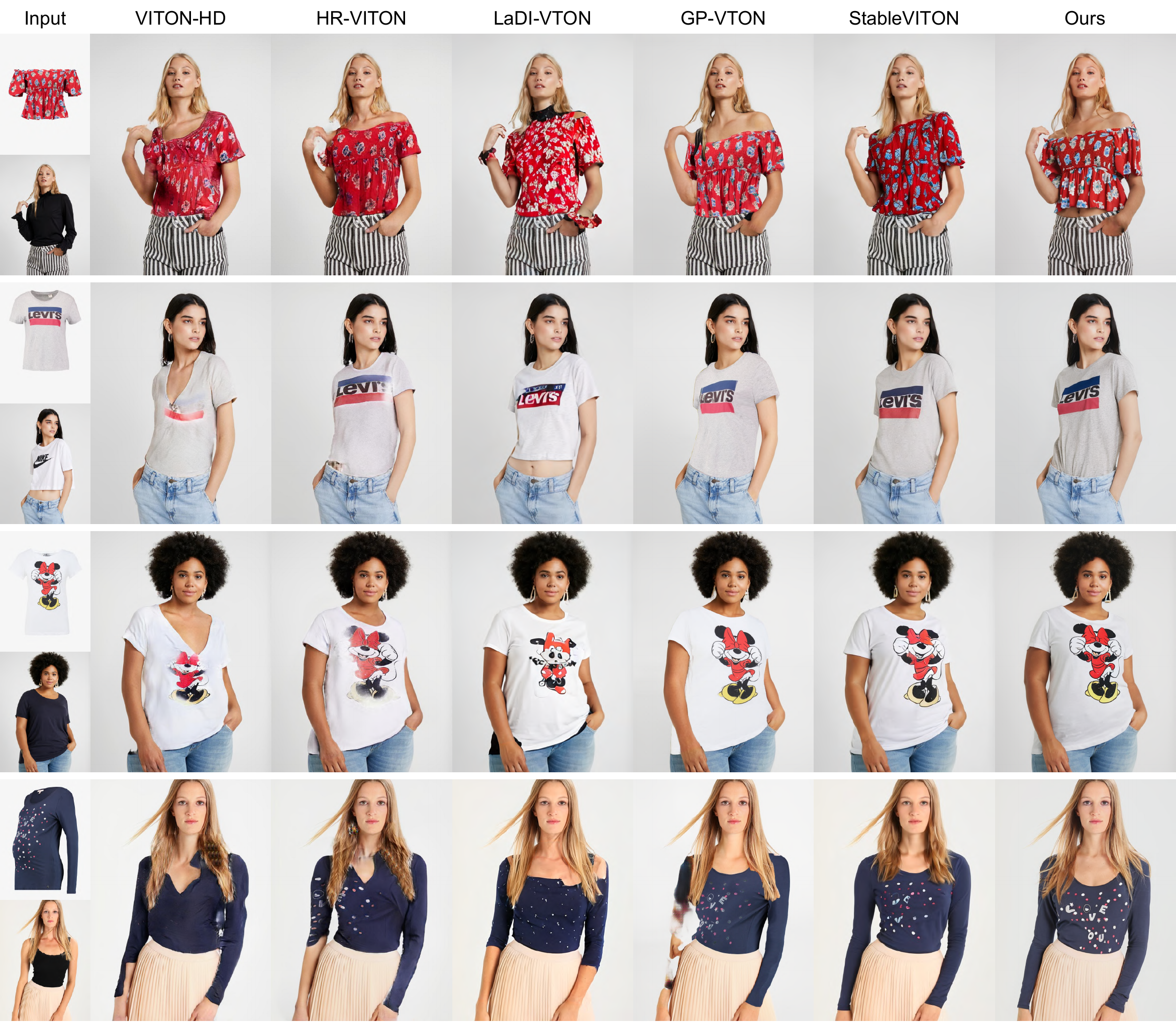}
  \caption{Qualitative comparison on the VITON-HD dataset~\cite{choi2021viton} (half-body models with upper-body garments). Please zoom in for more details.}
  \label{fig:hd}
  \vspace{-0.5cm}
\end{figure}

\begin{figure}[!t]
  \centering
  \includegraphics[width=\linewidth]{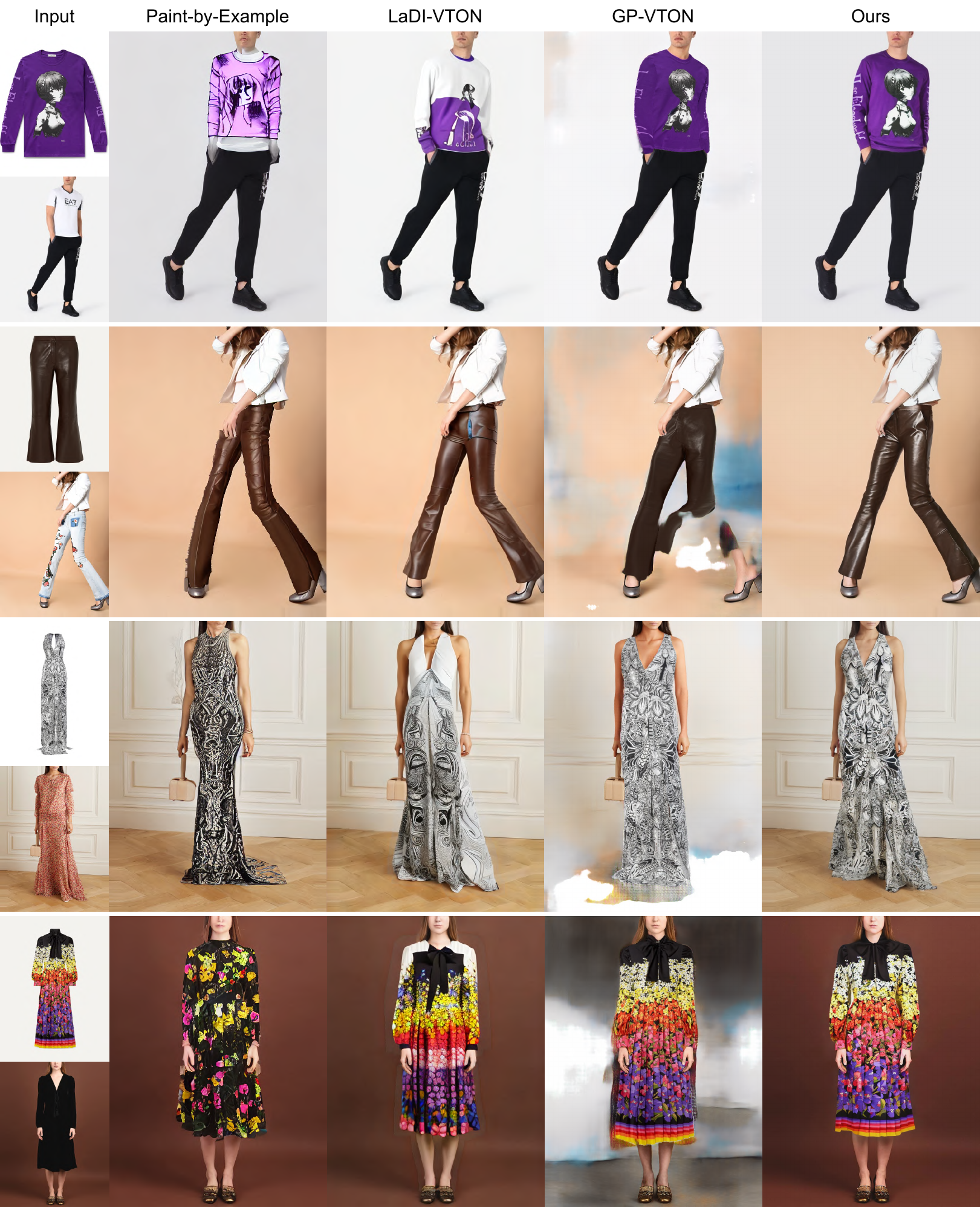}
  \caption{Qualitative comparison on the Dress Code dataset~\cite{morelli2022dress} (full-body models with upper-body garments/lower-body garments/dresses). Please zoom in for more details.}
  \label{fig:dc}
  \vspace{-0.5cm}
\end{figure}

\begin{figure}[!t]
  \centering
  \includegraphics[width=\linewidth]{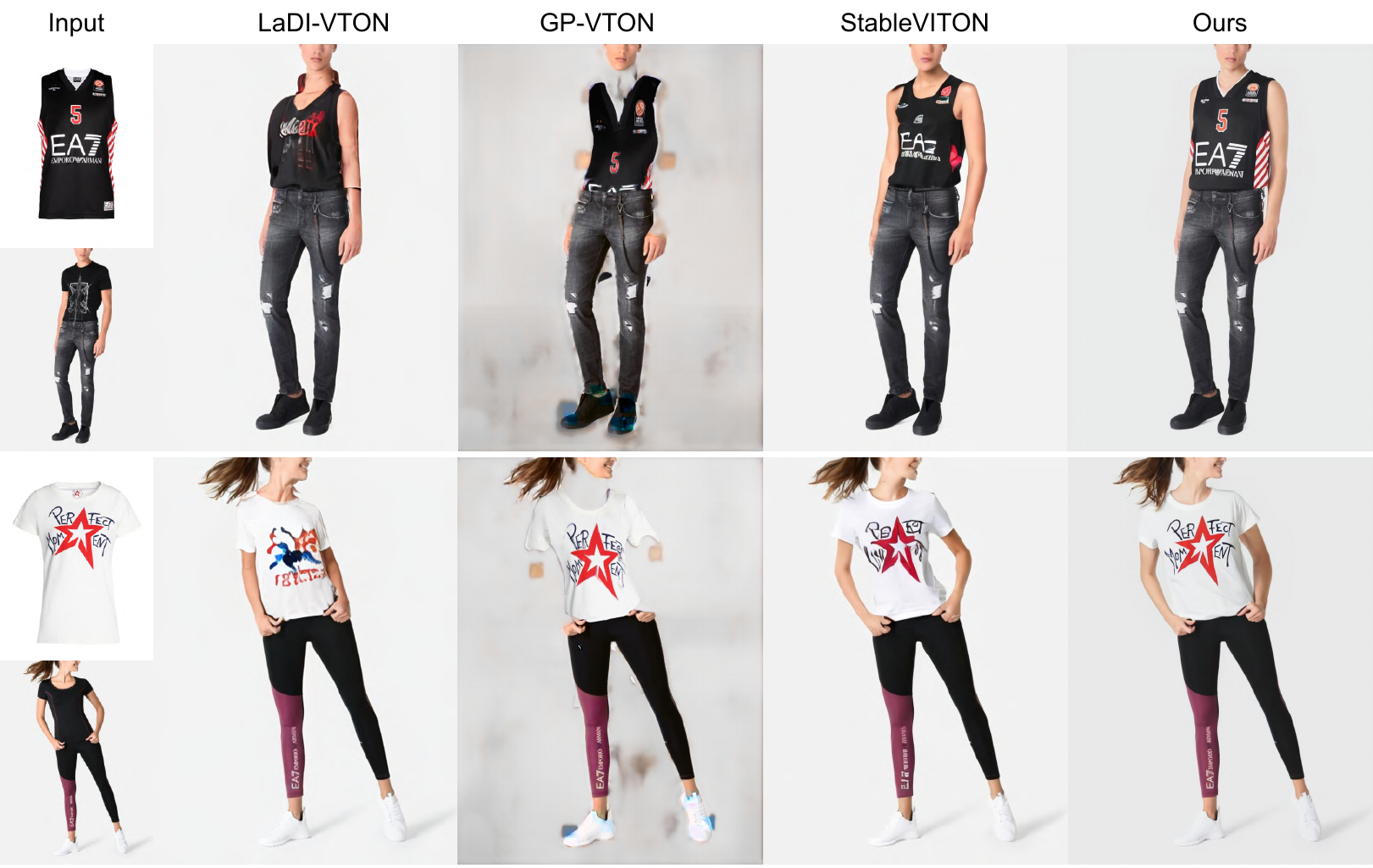}
  \caption{Qualitative results of the cross-dataset evaluation. The models are trained on the VITON-HD dataset~\cite{choi2021viton} and tested on the Dress Code dataset~\cite{morelli2022dress}. Please zoom in for more details.}
  \label{fig:crs}
  \vspace{-0.5cm}
\end{figure}

\subsection{Experimental Results}

\subsubsection{Qualitative Results.}
\cref{fig:hd} visually shows some example results of our method and other VTON methods on the test set of VITON-HD~\cite{choi2021viton}. We observe that compared with other methods, our OOTDiffusion consistently achieves the best try-on effects for various upper-body garments. More specifically, GAN-based methods like GP-VTON~\cite{xie2023gp} often fail to generate realistic human bodies (1st and 4th rows) or natural garment folds (2nd and 3rd rows), making the outfitted images look unrealistic. While other LDM-based methods including LaDI-VTON~\cite{morelli2023ladi} and StableVITON~\cite{kim2023stableviton} tend to lose some garment details such as complicated text (2nd and 4th rows) or patterns (1st and 3rd rows). In contrast, our OOTDiffusion not only generates realistic images but also preserves most of the fine-grained garment details.

Regarding the more complicated Dress Code dataset~\cite{morelli2022dress}, which consists of full-body models and various garment categories, our OOTDiffusion still visually outperforms other VTON methods. As illustrated in \cref{fig:dc}, Paint-by-Example~\cite{yang2023paint} and LaDI-VTON~\cite{morelli2023ladi} fail to preserve the garment features, and GP-VTON~\cite{xie2023gp} tends to cause severe body and background distortion. On the contrary, our OOTDiffusion consistently shows very stable performance on different garment categories including upper-body garments (1st row), lower-body garments (2nd row) and dresses (3rd and 4th rows).

In order to evaluate the generalization ability of our method, we conduct an additional cross-dataset experiment, i.e., training on one dataset and testing on the other. \cref{fig:crs} demonstrates that among all the models trained on the VITON-HD dataset~\cite{choi2021viton}, our OOTDiffusion is optimally adapted to the test examples in the Dress Code dataset~\cite{morelli2022dress}, generating more realistic outfitted images and preserving much more garment details.
In summary, the observations above (\cref{fig:hd,fig:dc,fig:crs}) qualitatively prove the superiority and generalization capability of our OOTDiffusion in generating natural and accurate try-on results for various human and garment images.

\subsubsection{Quantitative Results.}

\begin{table}[!t]
  \caption{Quantitative results on the VITON-HD dataset~\cite{choi2021viton}. The best and second best results are reported in \textbf{bold} and \underline{underline}, respectively.}
  \label{tab:vitonhd}
  \centering
    \begin{tabular}{lc cc cc}
    \toprule
    \textbf{Method} & & \textbf{LPIPS} $\downarrow$ & \textbf{SSIM} $\uparrow$ & \textbf{FID} $\downarrow$ & \textbf{KID} $\downarrow$\\
    \midrule
    VITON-HD~\cite{choi2021viton}         & & 0.116 & 0.863          & 12.13  & 3.22\\
    HR-VITON~\cite{lee2022high}           & & 0.097 & \underline{0.878}          & 12.30  & 3.82\\
    LaDI-VTON~\cite{morelli2023ladi}      & & 0.091 & 0.875          & 9.31   & 1.53\\
    GP-VTON~\cite{xie2023gp}              & & \underline{0.083} & \textbf{0.892} & 9.17   & \underline{0.93} \\
    StableVITON~\cite{kim2023stableviton} & & 0.084 & 0.862          & \underline{9.13}   & 1.20 \\
    \midrule
    \textbf{OOTDiffusion (Ours)}          & &\textbf{0.071} & \underline{0.878} & \textbf{8.81} & \textbf{0.82}\\
    \bottomrule
    \end{tabular}
\end{table}

\begin{table}[!t]
  \caption{Quantitative results on the Dress Code dataset~\cite{morelli2022dress}. The best and second best results are reported in \textbf{bold} and \underline{underline}, respectively. The * marker refers to the results reported in previous work.}
  \label{tab:dresscode}
  \centering
    \resizebox{\linewidth}{!}{
    \begin{tabular}{lc cccc c cc c cc c cc}
    \toprule
    \multirow{2}{*}{\textbf{Method}} & & \multicolumn{4}{c}{\textbf{All}} & & \multicolumn{2}{c}{\textbf{Upper-body}} & & \multicolumn{2}{c}{\textbf{Lower-body}} & & \multicolumn{2}{c}{\textbf{Dresses}}\\
    \cmidrule{3-6} \cmidrule{8-9} \cmidrule{11-12} \cmidrule{14-15}
     & & \textbf{LPIPS} $\downarrow$ & \textbf{SSIM} $\uparrow$ & \textbf{FID} $\downarrow$ & \textbf{KID} $\downarrow$ & &
    \textbf{FID} $\downarrow$ & \textbf{KID} $\downarrow$ & & \textbf{FID} $\downarrow$ & \textbf{KID} $\downarrow$ & & \textbf{FID} $\downarrow$ & \textbf{KID} $\downarrow$ \\
    \midrule
    PSAD*~\cite{morelli2022dress} & & 0.058 & 0.918 & 10.61 & 6.17 && 17.51 & 7.15 && 19.68 & 8.90 && 17.07 & 6.66\\
    Paint-by-Example~\cite{yang2023paint} & & 0.142 & 0.851 & 9.57 & 3.63 && 18.63 & 4.81 && 15.89 & 4.12 && 19.15 & 5.88\\
    LaDI-VTON~\cite{morelli2023ladi} & & 0.067 & 0.910 & \underline{5.66} & \underline{1.21} && 12.30 & 1.30 && \underline{13.38} & \underline{1.98} && 13.12 & 1.85\\
    GP-VTON~\cite{xie2023gp} & & \underline{0.051} & \underline{0.921} & 5.88 & 1.28 && \underline{12.20} & \underline{1.22} && 16.65 & 2.86 && \underline{12.65} & \underline{1.84}\\
    \midrule
    \textbf{OOTDiffusion (Ours)} & & \textbf{0.045} & \textbf{0.927} & \textbf{4.20} & \textbf{0.37} && \textbf{11.03} & \textbf{0.29} && \textbf{9.72} & \textbf{0.64} && \textbf{10.65} & \textbf{0.54}\\
    \bottomrule
    \end{tabular}
    }
\end{table}

\begin{table}[!t]
  \caption{Quantitative results of the cross-dataset evaluation. Each model is trained on one of the VITON-HD~\cite{choi2021viton} and Dress Code~\cite{morelli2022dress} datasets, and evaluated on the other. The best and second best results are reported in \textbf{bold} and \underline{underline}, respectively. The * marker refers to the results reported in previous work.}
  \label{tab:crossval}
  \centering
    \resizebox{\linewidth}{!}{
    \begin{tabular}{lc cccc c cccc}
    \toprule
    \textbf{Train/Test}& & \multicolumn{4}{c}{\textbf{VITON-HD/Dress Code}} & & \multicolumn{4}{c}{\textbf{Dress Code/VITON-HD}}\\
    \cmidrule{3-6} \cmidrule{8-11}
    \textbf{Method} & & \textbf{LPIPS} $\downarrow$ & \textbf{SSIM} $\uparrow$ & \textbf{FID} $\downarrow$ & \textbf{KID} $\downarrow$ & & \textbf{LPIPS} $\downarrow$ & \textbf{SSIM} $\uparrow$ & \textbf{FID} $\downarrow$ & \textbf{KID} $\downarrow$\\
    \midrule
    VITON-HD*~\cite{choi2021viton} & & 0.187 & 0.853 & 44.26 & 28.82 && - & - & - & -\\
    HR-VITON*~\cite{lee2022high} & & 0.108 & 0.909 & 19.97 & 7.35 && - & - & - & -\\
    LaDI-VTON~\cite{morelli2023ladi} & & 0.154 & 0.908 & 14.58 & 3.59 && \underline{0.235} & \underline{0.812} & \underline{29.66} & \underline{20.58}\\
    GP-VTON~\cite{xie2023gp} & & 0.291 & 0.820 & 74.36 & 80.49 && 0.266 & 0.811 & 52.69 & 49.14\\
    StableVITON~\cite{kim2023stableviton} & & \underline{0.065} & \underline{0.914} & \underline{13.18} & \underline{2.26} && - & - & - & -\\
    \midrule
    \textbf{OOTDiffusion (Ours)} & & \textbf{0.061} & \textbf{0.915} & \textbf{11.96} & \textbf{1.21} && \textbf{0.123} & \textbf{0.839} & \textbf{11.22} & \textbf{2.72}\\
    \bottomrule
    \end{tabular}
    }
    \vspace{-0.25cm}
\end{table}

\cref{tab:vitonhd} presents the quantitative evaluation results on the VITON-HD dataset~\cite{choi2021viton}. We find that some GAN-based models like HR-VITON~\cite{lee2022high} and GP-VTON~\cite{xie2023gp} achieve relatively high SSIM scores, indicating that they are able to retain the structural information of the original images. However, their generated images lack detail fidelity, and thus drop behind ours on LPIPS. The previous LDM-based methods including LaDI-VTON~\cite{morelli2023ladi} and StableVITON~\cite{kim2023stableviton} generate more realistic images according to their FID and KID scores, but they fail to restore the detail features due to their lossy feature fusion. In comparison, our OOTDiffusion not only generates realistic outfitted images but also preserves the precise details, and thus substantially outperforms other methods on the other three metrics (LPIPS, FID and KID) while obtaining comparable SSIM scores to the GAN-based methods.

\cref{tab:dresscode} demonstrates the state-of-the-art performance of our method on the Dress Code dataset~\cite{morelli2022dress}, which outperforms others on all the metrics for all the garment categories (upper-body/lower-body/dresses), confirming our feasibility in more complicated cases. Note that GP-VTON~\cite{xie2023gp} applies extra data modifications such as background removal and pose normalization to Dress Code, and only provides part of their test data. Despite this, our OOTDiffusion still achieves the best results on the more challenging original test dataset.

Furthermore, the generalization capability of our method is quantitatively verified by the results of the cross-dataset evaluation listed in \cref{tab:crossval}. We find that GP-VTON~\cite{xie2023gp} falls far behind other methods on all the metrics since its warping module severely overfits the training data. While our method leads again on all the metrics for the out-of-distribution test dataset. Overall, the observations above (\cref{tab:vitonhd,tab:dresscode,tab:crossval}) further demonstrate that our OOTDiffusion significantly outperforms previous VTON methods in both realism and controllability in all kinds of scenarios and conditions.

\subsection{Limitations}
Despite the state-of-the-art performance achieved in the image-based virtual try-on task, limitations still exist in our OOTDiffusion which demand further improvement. First, since our models are trained on paired human and garment images, it may fail to get perfect results for cross-category virtual try-on, e.g., to put a T-shirt on a woman in a long dress, or to let a man in pants wear a skirt. This issue can be partially solved in the future by collecting datasets of each person wearing different clothes in the same pose. Another limitation is that some details in the original human image might be altered after virtual try-on, such as muscles, watches or tattoos, etc. The reason is that the relevant body area is masked and repainted by the diffusion model. Thus more practical pre- and post-processing methods are required for addressing such problems.

\section{Conclusion}
In this paper, we present OOTDiffusion, a novel LDM-based network architecture for image-based vitrual try-on. The proposed outfitting UNet efficiently learns the garment features and incorporates them into the denoising UNet via the proposed outfitting fusion process with negligible information loss. Classifier-free guidance for the garment features is enabled by the proposed outfitting dropout in training, which further enhances the controllability of our method. Extensive experiments on high-resolution datasets show our superiority over other VTON methods in both realism and controllability, indicating that our OOTDiffusion has broad application prospects for virtual try-on.

\section*{Acknowledgements}
We sincerely thank our colleagues including Yilan Ye, Bin Fu, Wei Du, Xuping Su, and Chi Zhang, etc., for kindly supporting and promoting our work. Special thanks to Minh-Duc Vo for his helpful advice.

%
%
\bibliographystyle{splncs04}

\end{document}